\documentclass[letterpaper, 10 pt, conference]{ieeeconf}  
\pdfoutput=1

\IEEEoverridecommandlockouts                              

\usepackage{amsmath,graphicx,subcaption}
\usepackage[outdir=figures/]{epstopdf}

\usepackage{booktabs}
\usepackage{import}

\title{\bf
A CNN Segmentation-Based Approach to Object Detection and Tracking in Ultrasound Scans with Application to the Vagus Nerve Detection*
}

\author{Abdullah F. Al-Battal$^{1}$,  Yan Gong$^{1}$, Lu Xu$^{1}$, Timothy Morton$^{1}$, Chen Du$^{1}$, Yifeng Bu $^{1}$ \\Imanuel R Lerman$^{1,2}$, Radhika Madhavan$^{3}$, Truong Q. Nguyen$^{1}$ 
\thanks{*This work was supported in part by the Biological Advanced Research and Development Authority (BARDA Contract: 7515011900038), National Institute of Health (NIH Contract: IK2RX002920) and the David and Janice Katz Neural Sensor Research Fund in Memory of Allen E. Wolf.  © 2021 IEEE. Personal use of this material is permitted. Permission from IEEE must be obtained for all other uses, in any current or future
media, including reprinting/republishing this material for advertising or promotional purposes, creating new collective works, for resale or
redistribution to servers or lists, or reuse of any copyrighted component of this work in other works.}
\thanks{$^{1}$Electrical and Computer Engineering Department, UC San Diego, La Jolla, California, USA}%
\thanks{$^{2}$School of Medicine, UC San Diego, La Jolla, California, USA}
\thanks{$^{3}$General Electric Global Research, Schenectady, New York, USA}
}

\begin{document}

\maketitle
\thispagestyle{empty}
\pagestyle{empty}

\begin{abstract}
Ultrasound scanning is essential in several medical diagnostic and therapeutic applications. It is used to visualize and analyze anatomical features and structures that influence treatment plans. However, it is both labor intensive, and its effectiveness is operator dependent.  Real-time accurate and robust automatic detection and tracking of anatomical structures while scanning would significantly impact diagnostic and therapeutic procedures to be consistent and efficient. In this paper, we propose a deep learning framework to automatically detect and track a specific anatomical target structure in ultrasound scans.  Our framework is designed to be accurate and robust across subjects and imaging devices, to operate in real-time, and to not require a large training set. It maintains a localization precision and recall higher than $90\%$ when trained on training sets that are as small as $20\%$ in size of the original training set. The framework backbone is a weakly trained segmentation neural network based on U-Net. We tested the framework on two different ultrasound datasets with the aim to detect and track the Vagus nerve, where it outperformed current state-of-the-art real-time object detection networks. 

{\textbf{\textit{Clinical Relevance}}}\textemdash The proposed approach provides an accurate method to detect and localize target anatomical structures in real-time, assisting sonographers during ultrasound scanning sessions by reducing diagnostic and detection errors and expediting the duration of scanning sessions.

\end{abstract}

\section{Introduction}
\label{sec:intro}

Ultrasound scanning is an important step in many medical diagnostic and therapeutic workflows due to its well established safety record, its ability to visualize differences among soft tissues, and portability  \cite{klibanov2015ultrasound, miller2012overview}. However, ultrasound scanning is labor intensive where a scanning session can take up to 30 minutes. Ultrasound scans are  also sonographer dependent, creating relatively high cross operator variability in accurate anatomical structure identification; novice sonographers demonstrate high diagnostic error rates at up to $52\%$ more than expert sonographers \cite{tegnander2006examiner}.  Ultrasound imaging is primarily used to image soft tissue, which is inherently compressible, creating additional within-subject image variability.  Despite these limitations, the trained clinician must carry out accurate and precise ultrasound scanning as it is critical for identification of targeted structures, as well as for precise and accurate therapy administration \cite{klibanov2015ultrasound}. An automatic framework tool that assists sonographers in detecting and localizing anatomical structures may radically improve reliable across-subjects scanning  for both novice and expert sonographers. 

\begin{figure}[t]
\vspace{3pt}
\hspace{-8pt}
\begin{minipage}[b]{.48\linewidth}
  \centering
  \centerline{\includegraphics[trim = {0.72cm 0.5cm 0.75cm 0.1cm}, width=4.0cm, clip]{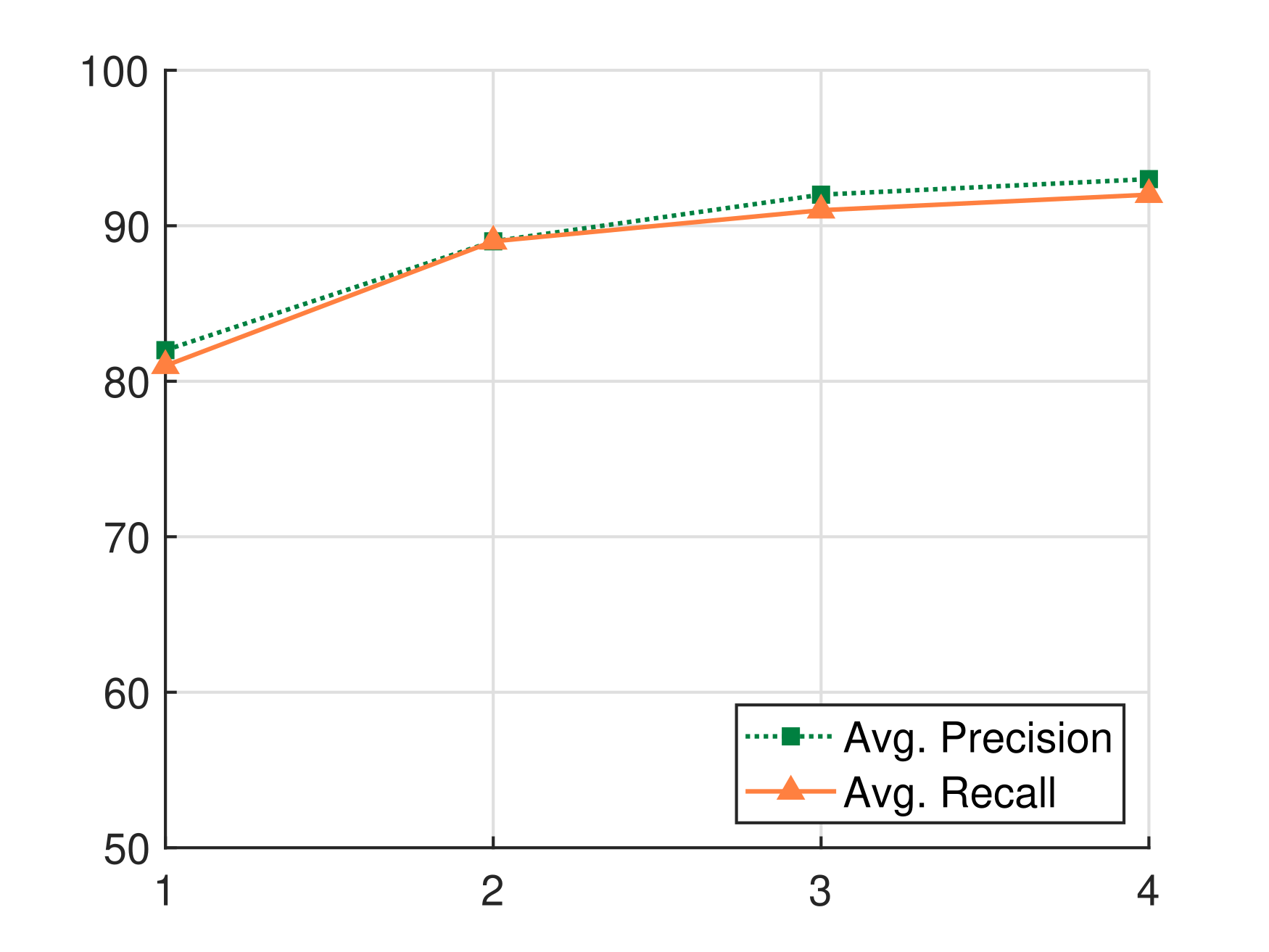}}
  \centerline{(a)}\medskip
\end{minipage}
\hfill
\begin{minipage}[b]{0.48\linewidth}
  \centering
  \centerline{\includegraphics[trim = {0.72cm 0.5cm 0.75cm 0.1cm}, width=4.0cm, clip]{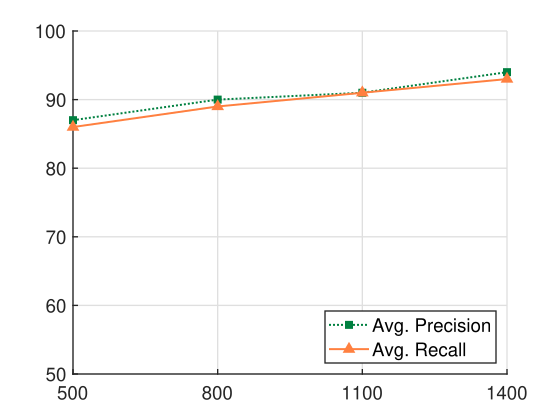}}
  \centerline{(b)}\medskip
\end{minipage}
\vspace{-0.27cm}
\caption{Object localization precision and recall for the 2\textsuperscript{nd}  dataset (a) versus the number of subjects used for training (testing on a separate subject), and (b) versus the number of training images (3 subjects for training and 2 for testing. Images chosen randomly from training set).
}
\label{fig:res1}
\vspace{-20pt}
\end{figure}

Object detectors are designed to localize objects and identify their underlying category or class within an image \cite{viola2001rapid, girshick2014rich}. Before the era of deep learning (DL), many traditional object detection algorithms used handcrafted features to detect objects that usually did not generalize well to real life situations. However, some prominent traditional methods, such as the Viola-Jones Detectors \cite{viola2001rapid}, Histogram of Oriented Gradients (HOG) \cite{dalal2005histograms}, and Deformable Part-based Model (DPM) \cite{felzenszwalb2008discriminatively} were quite successful. Nevertheless, since the introduction of DL based object detection algorithms, they outperformed traditional methods on every significant performance metric \cite{girshick2014rich, overfeat_2014}.

In medical imaging, object detection problems have been historically tackled using region of interest (ROI) tracking or segmentation-based approaches. For ROI tracking, several methods have been developed such as block matching \cite{Giachetti_2000} where exhaustive search-based block matching (ES-BM) is used to track anatomical structures such as arteries across sequential frames \cite{almekkawy2014two}, elliptical shape fitting to track and localize arteries and veins \cite{Wang_2009}, and deep learning methods using networks that compare similarities between frames \cite{Bharadwaj_2020_tracking}. Even though these ROI tracking methods have shown great potential in tracking objects in ultrasound scans, their ability to assist sonographers in detecting and localizing target anatomical structures during scanning sessions is hindered by their slow inference speeds \cite{Bharadwaj_2020_tracking, Bharadwaj_2020_faster}, or their dependency on operators to identify the target ROI at the beginning of a scanning session \cite{Giachetti_2000}, or both.

Segmentation-based approaches are used to recover a pixel-wise representation of every part within an image that belongs to an object \cite{u_net_2015}. Often, the goal of such algorithms is to identify the presence of an object in a medical scan, localize it, and estimate its size. These three goals are achievable through object detection algorithms \cite{overfeat_2014}, where the annotations used for training can be generated at a rate that is orders of magnitude faster than pixel-wise annotations used in segmentation algorithms.

\begin{figure*}[ht]
    \centering
    \includegraphics[width = 0.95\linewidth, height=6.5cm, trim = {0.25cm 0.25cm 0.25cm 0.25cm}, clip]{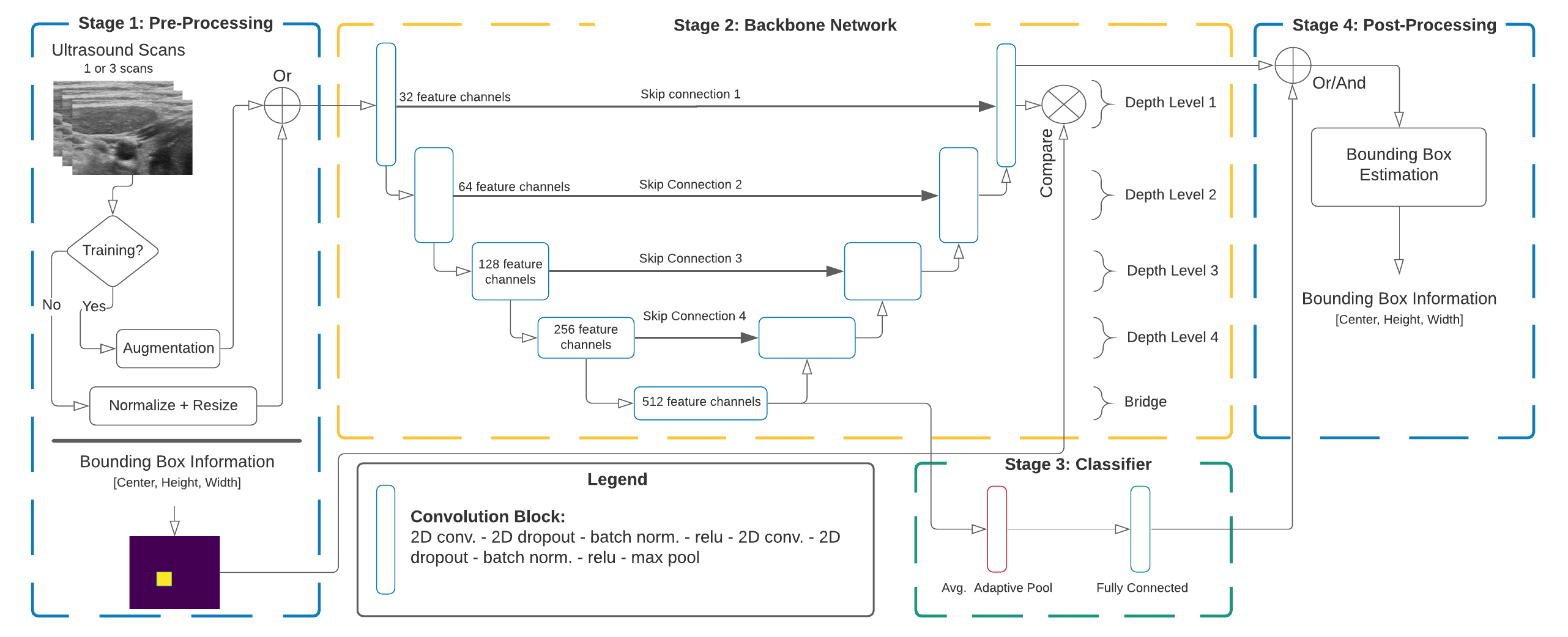}
    \caption{Overview of the proposed object detection framework with its four stages outlined.}
    \label{fig:sys}
    \vspace{-5mm}
\end{figure*}

In this paper, we propose a real-time object detection framework that is designed to autonomously detect, identify, and localize a specific anatomical structure in ultrasound scans. The specific anatomical structure we will identify is the cervical Vagus nerve encased within Carotid artery sheath.  The Carotid sheath encapsulated Vagus nerve sits at a variable depth (dependent on transducer probe to skin pressure) of approximately 1.2 to 2.5 cm (at high and low cervical transducer skin pressure due to jugular vein compression) \cite{lerman2016, mourdoukoutas2018high}. The proposed method uses a weakly supervised and modified U-Net convolutional neural network (CNN) as its backbone detection and localization algorithm \cite{u_net_2015}. It is designed to autonomously assist sonographers in real-time to enhance their ability to detect and track objects of interest during scanning sessions. We show that the proposed method outperforms YOLOv4 \cite{yolo_v4} and EffecientDet \cite{Efficient_Det}, the current state-of-the-art real-time object detection methods, in detecting the Vagus nerve.

\section{Related Work}
\label{sec:related}

\subsection{DL Based Object Detection}
\label{ssec:dl_obj_det}
Deep learning based object detection methods, and specifically CNN based methods, are currently the state-of-the-art \cite{overfeat_2014, girshick2014rich}. These detectors can be categorized into two broad categories, the two stage detectors such as Faster R-CNN \cite{ren2015faster}, and the one stage detectors such as SSD \cite{ssd_2016} and YOLO \cite{redmon2016you}.  Two stage detectors that defined the early success of DL based methods, are designed to have high identification and localization accuracy, while one stage detectors are designed to be fast and operate in real-time at 30+ frames per seconds (fps). Recently, these real-time methods achieved state-of-the-art object detection accuracy with a performance that is as good as, or better than, two stage methods \cite{Efficient_Det}.

However, both one and two stage detectors require large training sets. If one and two stage detectors are trained on smaller data sets (that is often times the case in medical imagery); overfiting and poor generalization can occur \cite{lee2018liver}.

\subsection{Object Detection in  Medical Imaging}
\label{ssec:med_obj_det}
Autonomous object detection in medical imaging has been historically treated as a segmentation problem. Pixel-wise annotations based on the presence of an object within an image are used to train detector algorithms to identify that object. In ultrasound, geometrical shapes fitting such as ellipse fitting for the detection of vessels, and contextual and textural features have been used to detect objects through a segmentation-based approach \cite{guerrero2007real, smistad2015real}.

 Most of the current advances in medical images' segmentation are based on DL approaches. One specific approach, the U-Net, that uses a contracting path followed by an expansive path, with skip connections between the two paths to preserve high resolution features localization, has proven to be highly trainable with a small training set making it very suitable for usage with medical images \cite{u_net_2015}. Several improvements have been made on the design of U-net, such as utilizing increased skip connections and deep supervision, have improved the accuracy performance of the network at the expense of segmentation time \cite{zhou2018unet++}.

Even though highly efficient in learning from smaller training sets when compared to object detectors, these segmentation approaches are expensive in their need for detailed annotations that require highly experienced medical personnel. Hence, a weakly supervised object detection framework that uses a modified U-Net trained as a backbone can achieve high accuracy with minimal supervision, reducing the cost of deployment while maintaining the capability to  accurately detect and localize target objects.


\section{Proposed Framework}
\label{sec:proposed}
Our proposed framework to detect, localize and track a specific target anatomical structure in real-time consists of 4 stages, as outlined in Fig. \ref{fig:sys}. The 1\textsuperscript{st} stage is designed to pre-process the scans, the 2\textsuperscript{nd} stage to detect and localize the target object within a scan, the 3\textsuperscript{rd} stage to classify whether a scan contains the target object, and the 4\textsuperscript{th} and final stage to fine tune the detection parameters. The framework is designed to have an inference latency of less than 33 ms when operating on a medium range graphical processing unit (GPU) such as the Nvidia RTX 2080 Ti.
\subsection{Stage 1: Pre-Processing}
\label{ssec:stage1}
In this stage, the frames are prepared for the backbone network in stages 2 and 3. The current ultrasound frame is stacked with the previous two frames as a three channel tensor of size $H\times W\times3$, where $H$ and $W$ are the height and width of the scan (frame). Using 3 frames instead of 1 has improved the localization accuracy by $1.7 \%$. The frames intensity values are then normalized to be in the range $[0, 1)$.

During training, we used extensive data augmentation to improve our framework's ability to overcome overfitting, and generalize to data outside of the training set \cite{NIPS2014_5548, wang2017effectiveness}. The augmentation pipeline includes geometrical transformations and color space randomized contrast and brightness transformations to account for differences in ultrasound signals' energy levels. Most importantly, the pipeline deployed:  1) deformable elastic transformations \cite{simard2003best} with random Gaussian kernels to elastically deform the grid of an image, which simulates the elastic differences among soft tissues within and across subjects, and 2) mixtures of input scans to enhance the coverage of the probability space while minimizing the risk function during training by implementing vicinal risk minimization instead of empirical risk minimization \cite{Chapelle_Vicinal_2000}. While computationally efficient, the empirical risk defined as $R_{e}(f) = 1/n \sum_{i=1}^{n}\ell(f(x_{i}),y_{i})$ only considers the performance of $f(x)$ on a finite set of training examples for a dataset consisting of training data $\mathcal{D} = \{(x_i,y_i)\}_{i=1}^{n} $ with $n$ examples of input ($x_i$) and target ($y_i$) pairs, prediction algorithm $f(x)$, and a loss function $\ell(f(x_{i}),y_{i})$. The empirical risk is used to approximate the expected risk, which is the average of the loss function $\ell$ over the joint distribution of inputs and targets $P(X,Y)$, where the joint distribution is only known at the training examples and can be approximated by the empirical distribution $P_{\delta}(x,y) = 1/n \sum_{i=1}^{n} \delta(x=x_i,y=y_i)$. However, the distribution can be approximated by $P_{v}(\tilde{x},\tilde{y}) = 1/n \sum_{i=1}^{n} \nu(\tilde{x},\tilde{y} | x_i,y_i)$ where $\nu$ is a vicinity distribution that computes the probability of finding the virtual input-target pair $(\tilde{x},\tilde{y})$ in the vicinity of the training input-target pair $(x_i,y_i)$ \cite{Chapelle_Vicinal_2000}. The virtual input-target pair $(\tilde{x},\tilde{y})$ can be defined as $\tilde{x} = \lambda x_i + (1-\lambda)x_j$ and $\tilde{y} = \lambda y_i + (1-\lambda)y_j$, where $(x_i,y_i)$ and $(x_j,y_j)$ are two randomly selected input-target pairs from the training set, and $\lambda$ is sampled from a beta distribution ($\lambda  \sim$ Beta$(\alpha, \alpha)$, $\alpha = 0.1$). This approximation offers a more comprehensive representation  and coverage of the joint distribution $P(X,Y)$. The use of mixtures of inputs have been implemented in image classification problems to minimize vicinal risk \cite{mixup2018}. In our framework, we have built and implemented an approach to use mixtures of inputs to minimize vicinal risk for segmentation-based algorithms.

In this stage, the masks that are used to train the network are created from bounding box coordinates as images (tensors) of size $H\times W$ where pixel values within the bounding box are set to 1. This mask will be used to weakly train the network in stage 2 to detect and localize the presence of target objects within the boundaries of the box.

\subsection{Stage 2: Backbone Detection Network}
\label{ssec:stage2}
Our framework's backbone network is designed based on a modified U-net architecture as shown in Fig. \ref{fig:sys}. The proposed network uses 4 depth levels as the standard U-Net with 2 convolutional layers in each depth level as well as the bridge of the network. However, in our proposed framework we used 32, 64, 128, 256, and 512 channels in the feature maps at levels 1, 2, 3, 4, and the bridge, respectively. The original method used twice the number of feature maps channels at each of these levels. Reducing the number of channels allows the network to operate in real-time. 

Reducing the size of a neural network usually reduces performance. To cope with this, we incorporated several modifications to improve the performance of the network such as two dimensional (2D) dropout layers in addition to original dropout layers. 2D dropout layers regularize the activations more efficiently when high correlation exists among pixels that are close to each other \cite{tompson2015efficient}. We also incorporated batch normalization layers and added a localization promoting term to the cost function. The original cost function of U-net is a confidence promoting loss function that computes the binary cross-entropy (BCE) between each pixel of the ground truth and predicted mask. For each element of the predicted mask with a value $x$ and ground truth value $y$ at location $(i,j)$, where $i = 1, 2, ..., H$ and $j = 1, 2, ..., W$, the BCE cost function can be computed for each training batch as:
\begin{equation}
\label{eq:bce}
\begin{gathered}
\mathcal{L}_{bce}(X,Y) = - \frac{1}{N} \sum_{n=1}^{N} \biggl[ \frac{1}{M} \sum_{m=1}^{M} \biggl[ \ell\left(x_{m,n} , y_{m,n}\right) \biggr] \biggr],
\end{gathered}
\end{equation}
\noindent where $N$ is the size of the batch, $M$ is the number of elements in the mask and is equal to $H \times W$,  $X$ is the predicted mask, $Y$ is the ground truth mask, and  $ \ell\left(x_{m,n} , y_{m,n}\right)$ is the loss computed element-wise between the ground truth and predictions, and is defined as: \begin{equation}
\label{eq:bce_element}
\begin{gathered}
\ell\left(x , y\right) =  w_{c} y \log{\sigma(x)} + (1-y) \log{\sigma(x)}.
\end{gathered}
\end{equation}
\noindent In (\ref{eq:bce_element}), the weight $w_{c}$ adjusts the loss function penalization for class $c$ based on the training set size imbalance for each class, and $\sigma(x)$ is the sigmoid function defined as $\sigma(x) = 1 / (1+exp(-x))$ and it maps the predicted elements into a probability space of predictions where $\sigma(x)$ constitutes an object if larger than or equal to 0.5, and background otherwise. The weight $w_{c}$ and the threshold of $\sigma(x)$ are used to influence the precision and recall of the network. The loss function promoting localization is based on the dice coefficient between the predicted and ground truth mask. The dice coefficient ($D_{c}$) is defined as \cite{dice_paper_segmentation}:

\begin{equation}
\label{eq:dice_coeff}
\begin{gathered}
D_{c}(\widehat{Y}, Y) =  \frac{2(\widehat{Y} \odot Y)}{\sum_{m=1}^{M}\widehat{y}_{m} + \sum_{m=1}^{M}y_{m}},
\end{gathered}
\end{equation}
\noindent where $\odot$ represents the element-wise multiplication and $\widehat{y}_{m} = \sigma(x_{m})$. The dice coefficient loss can then be defined to penalize lower $D_{c}$ values, which yields lower localization performance, as:
\begin{equation}
\label{eq:dice_loss}
\begin{gathered}
\mathcal{L}_{D_{c}}(X,Y) = 1 - D_{c}(\widehat{Y}, Y).
\end{gathered}
\end{equation}
\noindent The overall object detection loss function is defined as:
\begin{equation}
\label{eq:tot_loss}
\begin{gathered}
\mathcal{L}_{obj}(X,Y) = \alpha_{bce}\mathcal{L}_{bce}(X,Y) + \alpha_{dice}\mathcal{L}_{D_{c}}(X,Y),
\end{gathered}
\end{equation}
\noindent where $\alpha_{bce}$ and $\alpha_{dice}$ are coefficients that control the contribution of BCE loss and Dice loss, respectively, to the overall loss function. In our implementation we chose $\alpha_{bce} = 0.25$ and $\alpha_{dice} = 1$.

\subsection{Stage 3: Classifier}
\label{ssec:stage3}
Stage 2 is designed to localize an object within a scan, but is not optimized to identify the presence of the target object in the scan. Thus, to detect whether the target object is in the scan or not, we use a classifier optimized for this task as can be seen in Fig. \ref{fig:sys}. The classifier adds two extra layers to the framework and uses the output of the last layer of the bridge in stage 2, which contains 512 feature map channels, as input. This input is flattened to a tensor of length 512 using the average global pooling layer that was proposed as part of ResNet  \cite{he2016deep}. This is then followed by a fully-connected layer and an output layer for the 2 classes activated by a softmax function where the BCE loss (defined in (\ref{eq:bce})) is used to train the classifier.

\subsection{Stage 4: Post-Processing}
\label{ssec:stage4}

The output mask of the backbone network from stage 2 will be of size $H\times W$. After being threshold by the sigmoid function, the output mask will have elements with values between 0.5 and 1, as well as 0. These elements where the value is higher than 0.5, represent the region in which the network believes the target object is located. The average of the locations of these elements weighted by the confidence, which is the output of the sigmoid function $\sigma(x,y)$, is used to estimate the  center location of the target object. The center $(x_{c}, y_{c})$ of the target location can be then estimated as follows:
\begin{equation}
\label{eq:xc}
\begin{gathered}
x_{c} = \frac{\sum_{k=1}^{K} \sigma(x_{k})x_{k}}{\sum_{k=1}^{K} \sigma(x_{k})}, \;\;\;  y_{c} = \frac{\sum_{k=1}^{K} \sigma(y_{k})y_{k}}{\sum_{k=1}^{K} \sigma(y_{k})}.
\end{gathered}
\end{equation}

\noindent $K$ is the number of elements where the confidence $\sigma(x,y)$ is higher than the threshold, and $\sigma(x_{k}) = \sigma(y_{k}) = \sigma(x_{k}, y_{k})$. The weighted standard deviation of these elements' locations is used to estimate the width and height of the target object, and can be defined as follows for $x$:

\begin{equation}
\label{eq:x_std}
\begin{gathered}
\sigma_{x} = \sqrt{\frac{\sum_{k=1}^{K} \sigma(x_{k})(x_{k} - x_{c})^2}{\frac{(K-1)}{K}\sum_{k=1}^{K} \sigma(x_{k})}}, \\[4pt]
\end{gathered}
\end{equation}

\noindent where $\sigma_{x}$ is the standard deviation in the x direction. $\sigma_{y}$ can be calculated using (\ref{eq:x_std}) by replacing the corresponding variables. The width and height of the bounding box can then be calculated as: $width = \beta_{x}\sigma_{x}$ and $height = \beta_{y}\sigma_{y}$, where $\beta_{x}$ and $\beta_{y}$ are factors that are learned during the training of the backbone network. The output of the classifier is then fed together with the output of the backbone network to a decision logic such as an "or" or "and" to decide on the presence of the target object in the scan. Choosing "and" will increase precision at the expense of recall, and vice versa. Controlling this decision logic together with the thresholds ($\sigma(X)$) for the backbone network and classifier can be used in real-time by sonographers as simple methods to control the rate of false positives or false negatives.

\section{Experiments And Results}
\label{sec:Experiment}
\subsection{Dataset}
\label{ssec:dataset}
We evaluated our model on two different ultrasound datasets that were created by researchers at UC San Diego Health and Jacobs School of Engineering. The scans in the datasets were acquired to image the Vagus nerve in the mid-cervical and upper-cervical regions of the neck. The scans span different fields of view of the neck to create a variety of scans that would be generated by a sonographer who is looking to image the Vagus nerve within the neck. The two datasets were created using different probes and image reconstruction devices. The 1\textsuperscript{st} dataset was created using a probe and device with high quality diagnostic capabilities. The 2\textsuperscript{nd} dataset used a probe that has a small footprint to work alongside non-invasive therapeutic and stimulation devices, and is designed to generate scans at a very rapid pace at the expense of quality. The 1\textsuperscript{st} dataset contained 6,368 scans from 3 different subjects, while the 2\textsuperscript{nd} contained 26,313 scans from 5 different subjects. Both datasets contained scans from both the left and right side of the neck. The Vagus nerve shape, location and surrounding anatomical structures varies greatly within subjects and across subjects. Even a slight movement of the probe can make it  challenging for sonographers to re-identfy the nerve and its location due to the high variability of neck anatomy visualized with medio-lateral or cepahlo-caudal scanning along the cervical neck \cite{ahuja2019diagnostic}. In aggregate, nerve detection with variable anatomy datasets,  provides a substantial challenge for which we will test our proposed method and verify its robustness.

\subsection{Implementation and Setup}
\label{ssec:implementation}
We conducted 3 experiments to test the performance and robustness of our framework. The 1\textsuperscript{st} experiment was designed to test the accuracy of the proposed method in detecting and tracking the Vagus nerve on the 1\textsuperscript{st} dataset. The 2\textsuperscript{nd} experiment was designed to test the performance of the proposed method on the more challenging 2\textsuperscript{nd} dataset and compare it to YOLOv4 and EfficientDet, the current state-of-the-art real-time object detectors. In both of these experiments the datasets were divided into individual scans and split into a 64:16:20 ratio for training, validation, and testing, respectively. The 3\textsuperscript{rd} experiment was designed to test and verify the robustness of the framework in accounting for cross subject variabilities as well as its ability to generalize to new subjects. Hence, in this experiment, we divided our dataset by subject. The framework was trained on scans from 4 subjects and tested on the 5\textsuperscript{th} subject. Throughout all three experiments, the scans were resized to 256x256 for the 1\textsuperscript{st} dataset, and to 192x192 for the 2\textsuperscript{nd} dataset before being supplied to the network in stage 2. The backbone network was optimized using stochastic gradient descent (SGD) with a learning rate = $10^{-3}$, momentum = $0.9$, weight decay of $10^{-3}$, batch size = $16$, and trained for $200$ epochs. The proposed work was implemented in Pytorch  and our implementation is available in a Github repository \cite{vagus_nerve_u_net}.

\begin{table}[htb]
 \vspace{-1.25mm}
    \caption{The proposed framework localization precision and recall. For experiments 1 and 2, the threshold for a true positive is an IoU $\geq$ 0.5. For experiment 3, the threshold for a true positive is a distance error $\leq$ 2.5 mm from the nerve center.}
    \vspace{-1.75mm}
    \centering
    \begin{tabular*}{\linewidth}{l @{\extracolsep{\fill}} cc}
    \toprule
        Method & Avg. Precision & Avg. Recall \\
    \toprule
        Experiment 1 - $1^{st}$ Dataset\\
         \midrule
        \hspace{3mm}Ours - Single Frame & 94.4\%  & 97.2\%
        \vspace{1.75mm}
         \\
        Experiment 2  - $2^{nd}$ Dataset & & \\
    \midrule
        \hspace{3mm}Ours - Single Frame & 90.89\%  & 96.01\%  \\
        \hspace{3mm}Ours - Three Frames & \textbf{92.67\%}  & \textbf{97.29\%}  \\
        \hspace{3mm}YOLOv4 & 90.45\%  & 97.24\%  \\
        \hspace{3mm}EfficientDet - d3 & 91.93\%   & 96.35\%
        \vspace{1.75mm}
         \\

    Experiment 3 - $2^{nd}$ Dataset\\
    \midrule
    \hspace{3mm}Ours - Single Frame & 93.5\%  &  91.9\% \\
    \hspace{3mm}Ours - Three Frames & \textbf{95.1\%}  & \textbf{93.4\%} \\
    \bottomrule
    \end{tabular*}
    \label{tab:our_test}
    \vspace{-0.25cm}
\end{table}

\subsection{Evaluation and Results}
\label{ssec:evaluation}
To evaluate the accuracy and robustness of the proposed framework, we used the average precision and recall of localization, where a detection is considered a true positive when a certain localization threshold is met, otherwise that detection is considered a false positive. This is the main metric used to evaluate object detection algorithms \cite{pascal_2010}. For the 1\textsuperscript{st} and 2\textsuperscript{nd} experiments, the localization threshold is based on the intersection over union (IoU) metric. For the 3\textsuperscript{rd} dataset, the threshold is based on a physical distance of 2.5mm from the center of the nerve, which is equal to the radius of a typical Vagus nerve.

\begin{figure}[htb]
\vspace{2.5mm}
\begin{minipage}[b]{.48\linewidth}
  \centering
  \centerline{\includegraphics[width=3.8cm, height=3.2cm ]{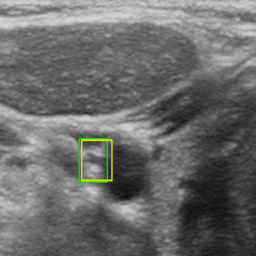}}
  \centerline{(a) 1\textsuperscript{st} dataset}\medskip
\end{minipage}
\hfill
\begin{minipage}[b]{0.48\linewidth}
  \centering
  \centerline{\includegraphics[width=3.8cm,  height=3.2cm]{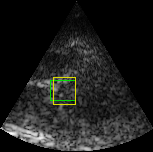}}
  \centerline{(b) 2\textsuperscript{nd} dataset}\medskip
\end{minipage}
\vspace{-3mm}
\caption{Ultrasound scans of the Vagus nerve with the ground truth (green) and predicted (yellow) bounding boxes.}
\vspace{-6mm}
\label{fig:res2}
\end{figure}

Table \ref{tab:our_test} summarizes the performance of the proposed framework for all of the 3 experiments. It can be seen that the proposed framework was able to identify and localize the Vagus nerve in both datasets with a high precision and recall. For the more challenging 2\textsuperscript{nd} dataset, the results of the 2\textsuperscript{nd} experiment show that the proposed method outperforms YOLOv4 and EfficientDet - d3. For the 3\textsuperscript{rd} experiment where the 2\textsuperscript{nd} dataset was divided by subjects, 4 subjects for training and 1 subject for testing, the proposed method was still able to generalize to subjects that it did not see during training and achieved high localization precision and recall. This was not the case for YOLOv4 and EffecientDet where the precision and recall dropped below $25\%$. This loss of accuracy can mainly be attributed to these methods' need for large training datasets with rich visual features to train their backbone and detection networks \cite{yolo_v4, Efficient_Det}. For the identification of target frames, the performance metrics for the classifier in stage 3 are 93.01\% for precision and 86.25\% for recall.

\begin{figure}[h!]
\vspace{-10pt}
\begin{center}
\includegraphics[trim = {1cm 0.2cm 0.9cm 0.4cm}, width=8.3cm,height=4.6cm, clip]{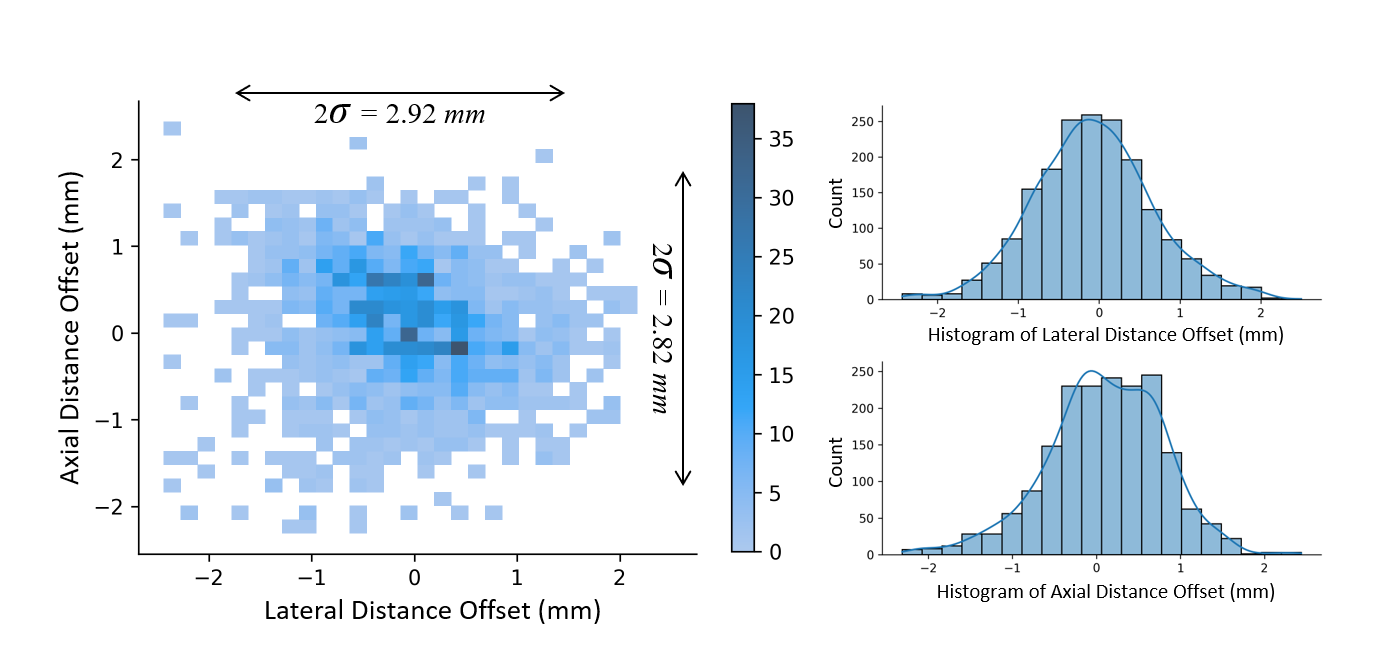}
\end{center}
\vspace{-0.27cm}
\caption{Heatmap (left) and histograms (right) of the lateral and axial distance offset of the true positive detections from the ground truth in millimeters.
}
\label{fig:res_final}
\vspace{-16pt}
\end{figure}

Fig. \ref{fig:res2} shows an example scan from each dataset with the ground truth and predicted bounding boxes. To verify the robustness of the proposed framework, we conducted two additional experiments to analyze the proposed framework performance on new subjects while being trained on smaller subsets of the original dataset.  The results of these two experiments are shown in Fig. \ref{fig:res1}. In the first experiment, we trained the framework on 1, 2, 3, and 4 subjects then tested on a 5\textsuperscript{th} subject and repeated this analysis twice for two different test subjects. We then used three subjects for training and two subjects for testing, randomly sampled scans from the training set, and created training subsets of sizes 500, 800, 110, and 1400. As observed in Fig. \ref{fig:res1}, the proposed framework has a high level of consistency and accuracy even when trained on smaller training sets. The framework produces high localization precision where more than 95$\%$ of the true positives predictions are located within 1.5 mm from the ground truth in both the lateral and axial directions, which is shown in the heat map and histograms of the true positive detections offset from the ground truth in Fig. \ref{fig:res_final}.

\section{Conclusion}
\label{sec:Conclusion}
We presented a weakly trained segmentation-based deep learning framework for real-time object detection and localization in ultrasound scans and tested its performance on detecting the Vagus nerve with an inference time of less than 33 ms. The framework used masks with bounding boxes enclosing the Vagus nerve as a target for the segmentation backbone network. It demonstrated that it can detect the Vagus nerve and localize it successfully with a limited number of training examples and without the need for time consuming and expensive pixel-wise  annotations, such as those needed for segmentation tasks.

\bibliographystyle{IEEEbib}
\bibliography{main}

\begin{thebibliography}{10}

\bibitem{klibanov2015ultrasound}
A.~L. Klibanov and J.~A. Hossack,
\newblock ``Ultrasound in radiology: from anatomic, functional, molecular
  imaging to drug delivery and image-guided therapy,''
\newblock {\em Investigative radiology}, vol. 50, no. 9, pp. 657, 2015.

\bibitem{miller2012overview}
D.~L. Miller, N.~B. Smith, M.~R. Bailey, \textit{et~al.},
\newblock ``Overview of therapeutic ultrasound applications and safety
  considerations,''
\newblock {\em Journal of ultrasound in medicine}, vol. 31, no. 4, pp.
  623--634, 2012.

\bibitem{tegnander2006examiner}
E.~Tegnander and S.~Eik-Nes,
\newblock ``The examiner's ultrasound experience has a significant impact on
  the detection rate of congenital heart defects at the second-trimester fetal
  examination,''
\newblock {\em Ultrasound in Obstetrics and Gynecology}, vol. 28, no. 1, pp.
  8--14, 2006.

\bibitem{viola2001rapid}
P.~Viola and M.~Jones,
\newblock ``Rapid object detection using a boosted cascade of simple
  features,''
\newblock in {\em Proceedings of the 2001 IEEE computer society conference on
  computer vision and pattern recognition. CVPR 2001}. IEEE, 2001, vol.~1, pp.
  I--I.

\bibitem{girshick2014rich}
R.~Girshick, J.~Donahue, T.~Darrell, and J.~Malik,
\newblock ``Rich feature hierarchies for accurate object detection and semantic
  segmentation,''
\newblock in {\em Proc. of the IEEE conference on computer vision and pattern
  recognition}, 2014, pp. 580--587.

\bibitem{dalal2005histograms}
N.~Dalal and B.~Triggs,
\newblock ``Histograms of oriented gradients for human detection,''
\newblock in {\em 2005 IEEE computer society conference on computer vision and
  pattern recognition}. IEEE, 2005, vol.~1, pp. 886--893.

\bibitem{felzenszwalb2008discriminatively}
P.~Felzenszwalb, D.~McAllester, and D.~Ramanan,
\newblock ``A discriminatively trained, multiscale, deformable part model,''
\newblock in {\em 2008 IEEE conference on computer vision and pattern
  recognition}. IEEE, 2008, pp. 1--8.

\bibitem{overfeat_2014}
P.~Sermanet, D.~Eigen, X.~Zhang, \textit{et~al.},
\newblock ``Overfeat: Integrated recognition, localization and detection using
  convolutional networks,''
\newblock in {\em 2nd International Conference on Learning Representations,
  {ICLR} 2014, Banff, AB, Canada, April 14-16, 2014}, 2014.

\bibitem{Giachetti_2000}
A.~Giachetti,
\newblock ``Matching techniques to compute image motion,''
\newblock {\em Image and Vision Computing}, vol. 18, no. 3, pp. 247--260, 2000.

\bibitem{almekkawy2014two}
M.~K. Almekkawy, Y.~Adibi, F.~Zheng, \textit{et~al.},
\newblock ``Two-dimensional speckle tracking using zero phase crossing with
  riesz transform,''
\newblock in {\em Proceedings of Meetings on Acoustics}. Acoustical Society of
  America, 2014, vol.~22.

\bibitem{Wang_2009}
D.~C. Wang, R.~Klatzky, B.~Wu, G.~Weller, A.~R. Sampson, and G.~D. Stetten,
\newblock ``Fully automated common carotid artery and internal jugular vein
  identification and tracking using b-mode ultrasound,''
\newblock {\em IEEE Transactions on Biomedical Engineering}, vol. 56, no. 6,
  pp. 1691--1699, 2009.

\bibitem{Bharadwaj_2020_tracking}
S.~{Bharadwaj} and M.~{Almekkawy},
\newblock ``Deep learning based motion tracking of ultrasound image
  sequences,''
\newblock in {\em 2020 IEEE International Ultrasonics Symposium (IUS)}, 2020,
  pp. 1--4.

\bibitem{Bharadwaj_2020_faster}
S.~{Bharadwaj} and M.~{Almekkawy},
\newblock ``Faster search algorithm for speckle tracking in ultrasound
  images,''
\newblock in {\em 2020 42nd Annual International Conference of the IEEE
  Engineering in Medicine Biology Society (EMBC)}, 2020, pp. 2142--2146.

\bibitem{u_net_2015}
O.~Ronneberger, P.~Fischer, and T.~Brox,
\newblock ``U-net: Convolutional networks for biomedical image segmentation,''
\newblock in {\em International Conference on Medical image computing and
  computer-assisted intervention}. Springer, 2015, pp. 234--241.

\bibitem{lerman2016}
I.~Lerman, R.~Hauger, L.~Sorkin, \textit{et~al.},
\newblock ``Noninvasive transcutaneous vagus nerve stimulation decreases whole
  blood culture-derived cytokines and chemokines: a randomized, blinded,
  healthy control pilot trial,''
\newblock {\em Neuromodulation: Technology at the Neural Interface}, vol. 19,
  no. 3, pp. 283--290, 2016.

\bibitem{mourdoukoutas2018high}
A.~P. Mourdoukoutas, D.~Q. Truong, D.~K. Adair, \textit{et~al.},
\newblock ``High-resolution multi-scale computational model for non-invasive
  cervical vagus nerve stimulation,''
\newblock {\em Neuromodulation: Technology at the Neural Interface}, vol. 21,
  no. 3, pp. 261--268, 2018.

\bibitem{yolo_v4}
A.~Bochkovskiy, C.~Wang, and H.~M. Liao,
\newblock ``Yolov4: Optimal speed and accuracy of object detection,''
\newblock {\em arXiv preprint arXiv:2004.10934}, 2020.

\bibitem{Efficient_Det}
M.~Tan, R.~Pang, and Q.~V. Le,
\newblock ``Efficientdet: Scalable and efficient object detection,''
\newblock in {\em Proc. of the IEEE/CVF Conference on Computer Vision and
  Pattern Recognition}, June 2020.

\bibitem{ren2015faster}
S.~Ren, K.~He, R.~Girshick, and J.~Sun,
\newblock ``Faster r-cnn: Towards real-time object detection with region
  proposal networks,''
\newblock in {\em Advances in neural information processing systems}, 2015, pp.
  91--99.

\bibitem{ssd_2016}
W.~Liu, D.~Anguelov, D.~Erhan, \textit{et~al.},
\newblock ``{SSD}: Single shot multibox detector,''
\newblock in {\em European conference on computer vision}. Springer, 2016, pp.
  21--37.

\bibitem{redmon2016you}
J.~Redmon, S.~Divvala, R.~Girshick, and A.~Farhadi,
\newblock ``You only look once: Unified, real-time object detection,''
\newblock in {\em Proceedings of the IEEE conference on computer vision and
  pattern recognition}, 2016, pp. 779--788.

\bibitem{lee2018liver}
S.~Lee, J.~S. Bae, H.~Kim, \textit{et~al.},
\newblock ``Liver lesion detection from weakly-labeled multi-phase ct volumes
  with a grouped single shot multibox detector,''
\newblock in {\em International Conference on Medical Image Computing and
  Computer-Assisted Intervention}. Springer, 2018, pp. 693--701.

\bibitem{guerrero2007real}
J.~Guerrero, S.~E. Salcudean, J.~A. McEwen, B.~A. Masri, and S.~Nicolaou,
\newblock ``Real-time vessel segmentation and tracking for ultrasound imaging
  applications,''
\newblock {\em IEEE transactions on medical imaging}, vol. 26, no. 8, pp.
  1079--1090, 2007.

\bibitem{smistad2015real}
E.~Smistad and F.~Lindseth,
\newblock ``Real-time automatic artery segmentation, reconstruction and
  registration for ultrasound-guided regional anaesthesia of the femoral
  nerve,''
\newblock {\em IEEE Transactions on Medical Imaging}, vol. 35, no. 3, pp.
  752--761, 2015.

\bibitem{zhou2018unet++}
Z.~Zhou, M.~M.~R. Siddiquee, N.~Tajbakhsh, and J.~Liang,
\newblock ``Unet++: A nested u-net architecture for medical image
  segmentation,''
\newblock in {\em Deep Learning in Medical Image Analysis and Multimodal
  Learning for Clinical Decision Support}, pp. 3--11. Springer, 2018.

\bibitem{NIPS2014_5548}
A.~Dosovitskiy, J.~T. Springenberg, M.~Riedmiller, and T.~Brox,
\newblock ``Discriminative unsupervised feature learning with convolutional
  neural networks,''
\newblock in {\em Advances in Neural Information Processing Systems 27}, pp.
  766--774. Curran Associates, Inc., 2014.

\bibitem{wang2017effectiveness}
J.~Wang and L.~Perez,
\newblock ``The effectiveness of data augmentation in image classification
  using deep learning,''
\newblock {\em Convolutional Neural Networks Vis. Recognit}, vol. 11, 2017.

\bibitem{simard2003best}
P.~Simard, D.~Steinkraus, and J.~Platt,
\newblock ``Best practices for convolutional neural networks applied to visual
  document analysis,''
\newblock in {\em Seventh International Conference on Document Analysis and
  Recognition, 2003. Proceedings.}, 2003, vol.~2, pp. 958--958.

\bibitem{Chapelle_Vicinal_2000}
O.~Chapelle, J.~Weston, L.~Bottou, and V.~Vapnik,
\newblock ``Vicinal risk minimization,''
\newblock in {\em Proc. of the 13th International Conference on Neural
  Information Processing Systems}, Cambridge, MA, USA, 2000, NIPS'00, p.
  395–401, MIT Press.

\bibitem{mixup2018}
H.~Zhang, M.~Cisse, Y.~N. Dauphin, and D.~Lopez-Paz,
\newblock ``mixup: Beyond empirical risk minimization,''
\newblock in {\em International Conference on Learning Representations}, 2018.

\bibitem{tompson2015efficient}
J.~Tompson, R.~Goroshin, A.~Jain, Y.~LeCun, and C.~Bregler,
\newblock ``Efficient object localization using convolutional networks,''
\newblock in {\em Proceedings of the IEEE conference on computer vision and
  pattern recognition}, 2015, pp. 648--656.

\bibitem{dice_paper_segmentation}
A.~P. {Zijdenbos}, B.~M. {Dawant}, R.~A. {Margolin}, and A.~C. {Palmer},
\newblock ``Morphometric analysis of white matter lesions in mr images: method
  and validation,''
\newblock {\em IEEE Transactions on Medical Imaging}, vol. 13, no. 4, pp.
  716--724, 1994.

\bibitem{he2016deep}
K.~He, X.~Zhang, S.~Ren, and J.~Sun,
\newblock ``Deep residual learning for image recognition,''
\newblock in {\em Proceedings of the IEEE conference on computer vision and
  pattern recognition}, 2016, pp. 770--778.

\bibitem{ahuja2019diagnostic}
A.~T. Ahuja,
\newblock {\em Diagnostic Ultrasound: Head and Neck},
\newblock Diagnostic Ultrasound. W. B. Saunders, 2019.

\bibitem{vagus_nerve_u_net}
Abdullah~F. Al-Battal,
\newblock ``Trbg/vagus-nerve-u-net,'' April 2021,
  http://github.com/trbg/vagus-nerve-u-net.

\bibitem{pascal_2010}
M.~Everingham, L.~Van~Gool, C.~K. Williams, \textit{et~al.},
\newblock ``The pascal visual object classes (voc) challenge,''
\newblock {\em International journal of computer vision}, vol. 88, no. 2, pp.
  303--338, 2010.

\end{thebibliography}

\end{document}